\documentclass[runningheads]{llncs}
\usepackage{graphicx}
\usepackage{hyperref}
    \hypersetup{
        colorlinks   = true,
        citecolor    = blue,
        linkcolor=blue  
    }
    
\usepackage{xstring}
\makeatletter
\AtBeginDocument{
\let\oldref\ref
\renewcommand{\ref}[1]{\IfBeginWith{#1}{fig:}%
{{\color{blue}Fig.~\oldref{#1}}}%
{\IfBeginWith{#1}{tab:}{{\color{blue}Table~\oldref{#1}}}%
{\IfBeginWith{#1}{sec:}{{\color{blue}Section~\oldref{#1}}}%
{\IfBeginWith{#1}{Eq:}{{\color{blue}Eq.~\oldref{#1}}}%
{\IfBeginWith{#1}{appendix1}{{\color{blue}Appendix 1}}}%
{}}}
}}}

\begin{document}
\title{A Genetic Algorithm based Kernel-size Selection Approach for a Multi-column Convolutional Neural Network}


\author{Animesh Singh\inst{1} \and
Sandip Saha\inst{2}\and
Ritesh Sarkhel \inst{3} Mahantapas  Kundu\inst{4}, Mita Nasipuri \inst{5}    \and
Nibaran Das\inst{6}}

\authorrunning{Animesh Singh et al.}

\institute{Jalpaiguri Govt. Engineering College, West Bengal, India\\
\email{as2002@cse.jgec.ac.in} \and
Jalpaiguri Govt. Engineering College, West Bengal, India\\
\email{ss2054@cse.jgec.ac.in}\\ \and
The Ohio State University, Ohio, Columbus\\
\email{sarkhel.5@osu.edu}\\ \and
Jadavpur University, West Bengal, India\\
\email{mahantapas.kundu@jadavpuruniversity.in}\\ \and
Jadavpur University, West Bengal, India\\
\email{mita.nasipuri@jadavpuruniversity.in}\\ \and
Jadavpur University, West Bengal, India\\
\email{nibaran.das@jadavpuruniversity.in}
}

\maketitle              
\begin{abstract}
Deep neural network based architectures give promising results in various domains including \textit{pattern recognition}. Finding the optimal combination of the hyper-parameters of such a large-sized architecture is tedious and requires large number of laboratory experiments. But, identifying the optimal combination of a hyper-parameter or appropriate kernel size for a given architecture of deep learning  is always challenging and tedious task. Here, we introduced a genetic algorithm based technique to reduce the efforts of finding the optimal combination of a hyper-parameter (kernel size) of a convolutional neural network based architecture. The method is evaluated on three popular datasets of different handwritten Bangla characters and digits.
The implementation of the proposed methodology can be found in the following link:
\url{https://github.com/DeepQn/GA-Based-Kernel-Size}.

\end{abstract}

\section{Introduction}
Deep neural network based architectures \cite{cirecsan2012multi,lecun2015deep,zhang2016single,sarkhel2019deterministic} have shown its efficacy in various challenging problems of multiple different domains throughout the last decade. Among those domains, handwritten characters or digits recognition is one of the important one. Researchers of this field have proposed numerous neural network based architectures to achieve better recognition accuracies. Researchers like Ciresan et al. \cite{cirecsan2012multi}, Sarkhel et al. \cite{sarkhel2017multi}, Gupta et al. \cite{gupta2019multiobjective}, Roy et al. \cite{roy2017handwritten}, Krizhevsky et al. \cite{krizhevsky2012imagenet} etc have proposed different neural network based architectures and optimization techniques for recognition of different handwritten characters and digits. Most of these researchers configured their proposed neural network based architectures empirically on the basis of exploratory laboratory experiments. Most recent researches \cite{zoph2016neural,pham2018efficient,lu2019nsga,liu2018progressive} show that the optimal structure of the neural network based architectures can be found with Reinforcement learning \cite{mnih2015human} based techniques. With the help of these techniques, the effort of rigorous laboratory experiment for finding the best structure of a neural network can be reduced to a significant extent.
\newline
\newline
In our present work we proposed a genetic algorithm \cite{das2012genetic} based approach to find the optimal structure of a multi-column convolutional neural network (CNN) \cite{cirecsan2012multi}. More specifically the proposed genetic algorithm based technique selects optimal combination of the kernel-sizes for the layers of our proposed multi-column CNN based architecture. Our proposed technique has been tested on three publicly available different datasets of Bangla scripts. The results showed a significant improvement, which proves the efficiency of the proposed methodology.
\newline
\newline
The rest of the paper is organized as follows: the detailed description of the proposed methodology is presented in \ref{sec:PM}, the training of the proposed architecture is described in \ref{sec:TA}, the experimental results are presented in \ref{sec:EX} and finally, a brief conclusion is drawn from the results in \ref{sec:CO}.

\begin{figure}
\centering \makeatletter\IfFileExists{Multi-Column.jpg}{\includegraphics[width=1\textwidth]{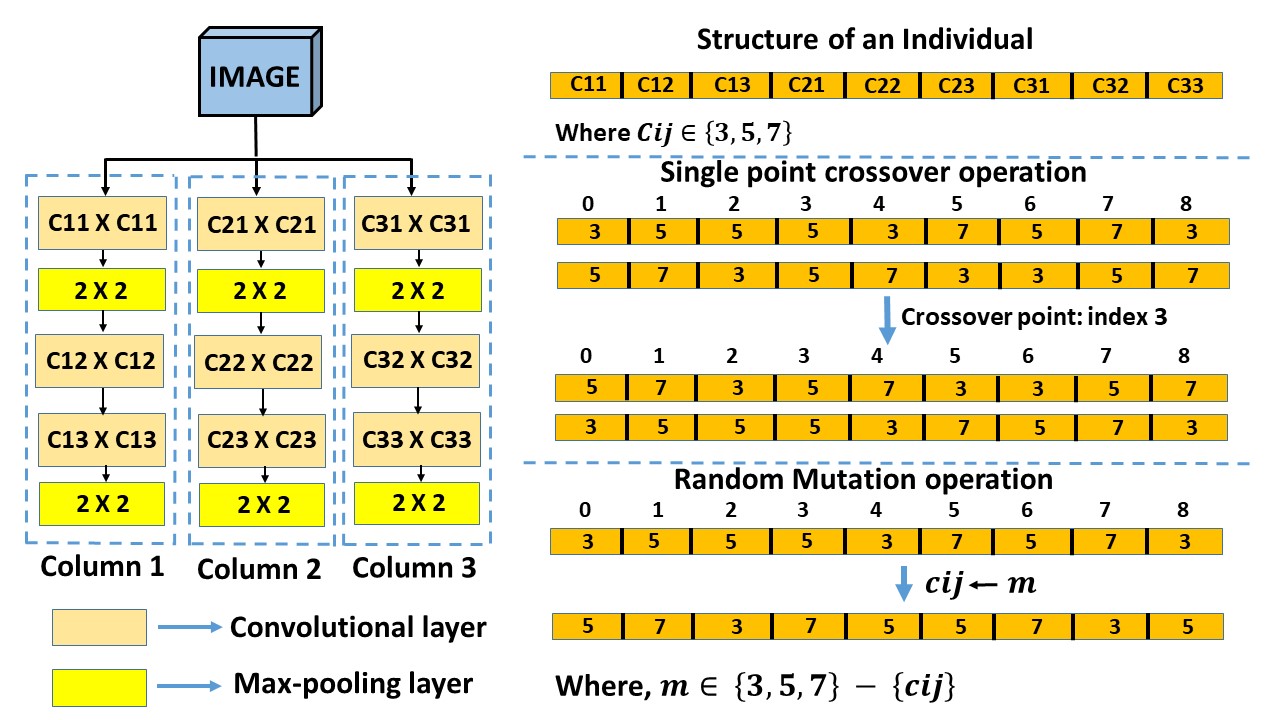}}{}
\makeatother 
\caption{{A high-level representation of different stages of our proposed \textit{genetic algorithm} based approach}}
\label{fig:GA}
\end{figure}

\section{Proposed Methodology} \label{sec:PM}
A detailed description of the proposed method is presented in this section. As mentioned earlier, a genetic algorithm based approach has been proposed for finding the optimal structure of a multi-column CNN used for recognizing handwritten characters and digits of multiple Bangla scripts. Generally, kernel sizes for the layers of a CNN architecture are finalized through experiments. Our proposed genetic algorithm based approach selects the optimal combination of kernel sizes for the different layers of every column of the multi-column CNN architecture.

\subsection{A brief overview of the proposed genetic algorithm based approach}
The primary stages of the proposed genetic algorithm based approach are described in the following:

\subsubsection{Population Initialization:}
The initial population of the algorithm is a collection of different sets of kernel combination (different sizes). That means each individual is a set of kernel sizes. The collection of different individuals are initialized randomly. The structure of an individual is a set containing the equal number of elements as the number of total layers of all the columns in the multi-column CNN architecture i.e. if each column has $L$ layers and there exists $C$ such columns then length of each individual will be $L \times C$ (shown in \ref{fig:GA}). The values of the elements will be $3$, $5$ or $7$ as only $3\times3$, $5\times5$ and $7\times7$ kernel sizes are used.

\subsubsection{Parent Selection:}
Individuals from the population are selected as parents which will undergo crossover and mutation operation to produce offspring with generally better fitness value than the parents. These parents are selected with Roulette Wheel parent selection \cite{lipowski2012roulette} mechanism.

\subsubsection{Crossover Operation:}
In our proposed genetic algorithm based approach single-point crossover operation is performed. In this operation two parent individuals participate and for each participating individual an index is randomly chosen and the elements of both the individuals are swapped against the randomly chosen index. That means, the elements in the left-hand part of the index of an individual are exchanged with the elements in the right-hand part of the index of another individual (shown in \ref{fig:GA}).

\subsubsection{Mutation Operation:}
A random mutation operation has been used in our proposed genetic algorithm. In this mutation operation the number of elements as well as the elements themselves are selected randomly from an individual. That means, a random number say $r$ is chosen between $0$ and length of an individual. Then randomly select $r$ different elements from the individual and replace those elements with different kernel sizes (chosen randomly excluding the existing kernel value at that chosen index location) as shown in \ref{fig:GA}.

\begin{figure}
\centering \makeatletter\IfFileExists{Model.jpg}{\includegraphics[width=1\textwidth]{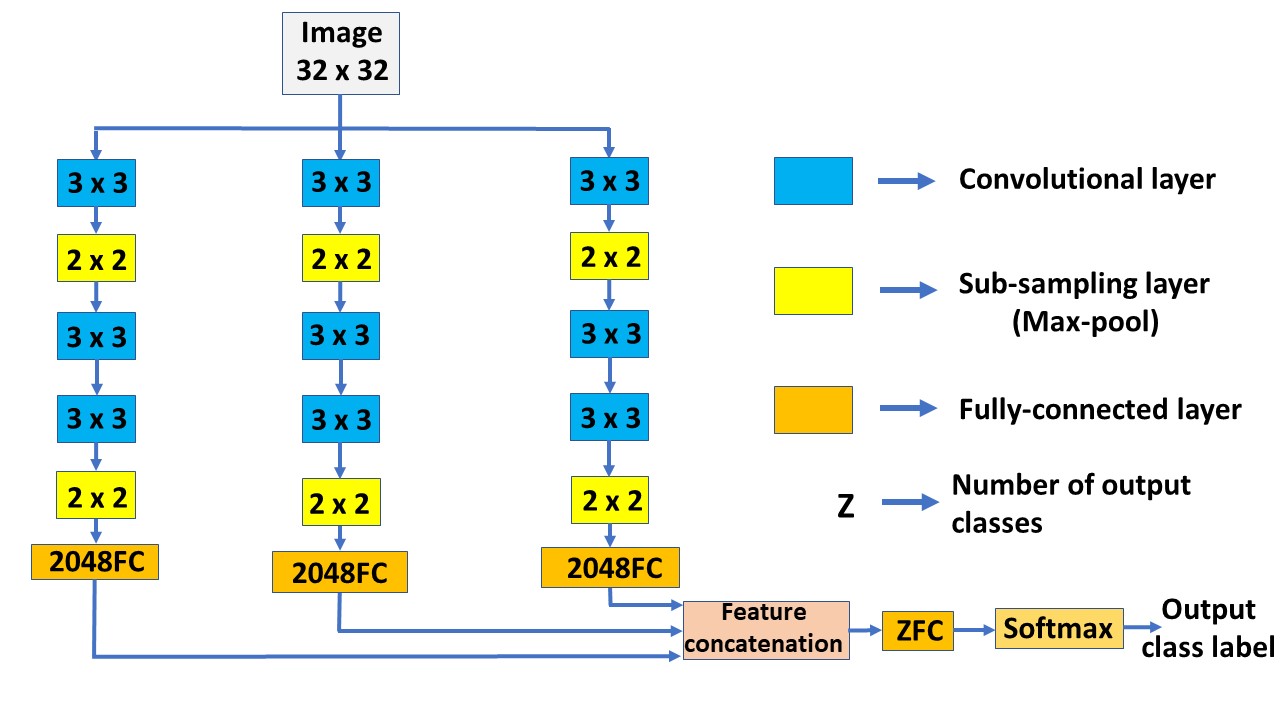}}{}
\makeatother 
\caption{{A graphical abstract of the initial configuration of the \textit{proposed multi-column CNN based architecture} used in our present work}}
\label{fig:Model}
\end{figure}

\begin{table}
\caption{Parameters of the proposed genetic algorithm} \label{tab:Genetic_Parameters}
\vspace{0.1cm}
\centering
\resizebox{0.5\columnwidth}{!}{%
\begin{tabular}{c|c}
\hline
Parameter name & Parameter value \\
\hline
 Population size & 100 \\ [0.1cm]
 \hline
 Maximum generation & 20 \\ [0.1cm]
 \hline
 Fitness value & Recognition accuracy on validation set \\ [0.1cm]
 \hline
 Parent selection method & Roulette wheel selection \\ [0.1cm]
 \hline
 Crossover type & Single point crossover \\ [0.1cm]
 \hline
 Mutation type & Random mutation \\ [0.1cm]
 \hline
 Crossover rate & 0.5 \\ [0.1cm]
 \hline
 Mutation Probability & 0.2 \\
\hline
\end{tabular}%
}
\end{table}

\begin{table}
\caption{Parameters of learning algorithm} \label{tab:Train_Parameters}
\vspace{0.1cm}
\centering
\resizebox{0.5\columnwidth}{!}{%
\begin{tabular}{c|c}
\hline 
Parameter name & Parameter value \\
\hline
 Learning algorithm & RMSProp \\ [0.1cm]
 \hline
 Initial learning rate & 0.001 \\ [0.1cm]
 \hline
 Learning decay rate & 0.05 \\ [0.1cm]
 \hline
 Dropout & 0.5 \\ [0.1cm]
 \hline
 Batch size & 250 \\ [0.1cm]
 \hline
 Training epochs per generation & 100 \\
\hline
\end{tabular}%
}
\end{table}

\begin{table}
\caption{Datasets used in the present work} \label{tab:Dataset}
\vspace{0.1cm}
\centering
\resizebox{\linewidth}{!}{%
\begin{tabular}{c|c|c|c|c|c}
\hline 
Index & Name of the dataset & Dataset type & Number of training samples & Number of test samples & Reference \\
\hline
 D1 & ISIBanglaDigit & Bangla Digits & 23,500 & 4000 & \cite{bhattacharya2008handwritten} \\ [0.2cm]

 D2 & CMATERdb 3.1.2 & Bangla basic characters & 12,000 & 3000 & \cite{sarkar2012cmaterdb1} \\ [0.2cm]

 D3 & CMATERdb 3.1.3 & Bangla compound characters & 34,229 & 8468 & \cite{alif2017isolated} \\

\hline
\end{tabular}%
}
\end{table}

\subsection{Architectures of the proposed multi-column CNN}
The initial configuration of every column of the multi-column CNN based architecture is following: 32C2-2P2-BN-RELU-128C1-BN-RELU-256C2-2P2-BN-RELU-2048FC. Some shorthand notations are used for space limitations. In this representation XCY denotes a convolutional layer with a total of X kernels and stride of Y pixels, MPN denotes a max-pooling \cite{nagi2011max} layer with an $M\times M$ pooling window and stride of N pixels, NFC denotes a fully-connected layer with N output neurons, BN denotes Batch-Normalization \cite{ioffe2015batch}, RELU denotes a Rectified Linear Unit activation layer \cite{dahl2013improving}. Finally, the output feature maps are combined with feature concatenation and passed through a Softmax classifier as following: ZFC-Softmax. Here Z denotes the total number of classes. A graphical view of initial configuration of the proposed architecture is given in \ref{fig:Model}.
\newline
\newline
The strides are not fixed for every kernel sizes. As kernel sizes vary during the optimal structure finding stage, the strides values are also varied accordingly. For kernel size $3\times3$ and $5\times5$ we have used a stride of $2$ pixels and for kernel size $7\times7$ we have used a stride of $1$ pixel.
\newline
\newline
As most of the information loss occurs at the max-pooling layers, only $2\times2$ kernel size is used in the subsampling layers. On the other hand, as the convolutional sampling layers are comparatively less problematic kernels with $3\times3$, $5\times5$ and $7\times7$ sizes are permissible \cite{sarkhel2017multi}.

\begin{table}
\caption{Recognition accuracy achieved by our proposed methodology on different datasets} \label{tab:Results}
\centering
\resizebox{\linewidth}{!}{%
\begin{tabular}{c|c|c|c|c}
\hline
Name of the dataset & Dataset type & Number of classes & Test set accuracy & Validation set accuracy \\
\hline
 ISIBanglaDigit & Bangla Digits & 10 & 99.12 \% & 99.78 \% \\ [0.1cm]

 CMATERdb 3.1.2 & Bangla basic characters & 50 & 97.10 \% & 96.57 \% \\ [0.1cm]

 CMATERdb 3.1.3 & Bangla compound characters & 171 & 94.77 \% & 94.88 \% \\

\hline
\end{tabular}%
}
\end{table}

\begin{table}
\caption{Recognition accuracy achieved by the proposed architecture on using $3 \times 3$ kernel size in every convolutional layer} \label{tab:Results3}
\centering
\resizebox{\linewidth}{!}{%
\begin{tabular}{c|c|c|c}
\hline
Name of the dataset & Dataset type & Number of classes & Test set accuracy \\
\hline
 
 ISIBanglaDigit & Bangla Digits & 10 & 98.98 \% \\ [0.1cm]

 CMATERdb 3.1.2 & Bangla basic characters & 50 & 95.45 \% \\ [0.1cm]

 CMATERdb 3.1.3 & Bangla compound characters & 171 & 92.56 \% \\

\hline
\end{tabular}%
}
\end{table}

\begin{table}
\caption{Recognition accuracy achieved by the proposed architecture on using $5 \times 5$ kernel size in every convolutional layer} \label{tab:Results5}
\centering
\resizebox{\linewidth}{!}{%
\begin{tabular}{c|c|c|c}
\hline
Name of the dataset & Dataset type & Number of classes & Test set accuracy \\
\hline
 
 ISIBanglaDigit & Bangla Digits & 10 & 99.02 \% \\ [0.1cm]

 CMATERdb 3.1.2 & Bangla basic characters & 50 & 95.56 \% \\ [0.1cm]

 CMATERdb 3.1.3 & Bangla compound characters & 171 & 93.12 \% \\

\hline
\end{tabular}%
}
\end{table}

\begin{table}
\caption{Recognition accuracy achieved by the proposed architecture on using $7 \times 7$ kernel size in every convolutional layer} \label{tab:Results7}
\centering
\resizebox{\linewidth}{!}{%
\begin{tabular}{c|c|c|c}
\hline
Name of the dataset & Dataset type & Number of classes & Test set accuracy \\
\hline
 
 ISIBanglaDigit & Bangla Digits & 10 & 98.87 \% \\ [0.1cm]

 CMATERdb 3.1.2 & Bangla basic characters & 50 & 95.10 \% \\ [0.1cm]

 CMATERdb 3.1.3 & Bangla compound characters & 171 & 91.46 \% \\

\hline
\end{tabular}%
}
\end{table}

\section{Training the Proposed Architecture} \label{sec:TA}
The proposed architecture is trained with every individual from the population over 20 generations. After every generation the best fit (with highest fitness value) individuals are selected for our proposed crossover and mutation operations. The best fit offspring is then used to update the population by replacing the individual of minimum fitness value. The parameters of the proposed genetic algorithm is presented in \ref{tab:Genetic_Parameters}.
\newline
\newline
The fitness value of each individual is found during training of our proposed architecture. The training images are fed into every column the architecture simultaneously and the loss function is calculated at the end Softmax classifier \cite{duan2003multi}. The connection weights between the layers of every column of our proposed architecture are updated in a single pass of the backpropagation after every epoch using equation \ref{Eq:Rmsprop1} and \ref{Eq:Rmsprop2}.

\begin{equation}\label{Eq:Rmsprop1}
E(g^2)_{t} = \beta E(g^2)_{t-1} + (1-\beta)(\frac{\partial C}{\partial w})^2
\end{equation}

\hspace{-0.4cm}where, $E[g]$ is moving average of squared gradients,  $\frac{\partial C}{\partial w}$ is gradient of the cost function with respect to the weight, $\eta$ learning rate, $\beta$ is moving average parameter.
\vspace{0.3cm}

\begin{equation}\label{Eq:Rmsprop2}
w_{t} = w_{t-1} - \frac{\eta}{\sqrt{E(g^2)_{t}}}\frac{\partial C}{\partial w}
\end{equation}
$w_{t}$ is value of the weight parameter at iteration t.
\newline
\newline
An adaptive learning rate RMSProp \cite{basu2018convergence} learning algorithm is used to train our proposed architecture. CrossentropyLoss \cite{farahnak2016multi} is used as the loss function (shown in \ref{Eq:CEL}) during training of the proposed architecture. In every generation the proposed architecture is trained for $100$ epochs with every individual of the population. A variable learning rate \cite{cirecsan2012multi} is used i.e. the learning rate is decreased by a factor of $0.05$/epoch until it reaches to the value of $0.00001$ while training. LeCun et al. \cite{lecun2012efficient} has suggested a technique of data shuffling before every epoch of training. As, shuffling introduces heterogeneity in the datasets and enhances the convergence rate of the learning algorithm, in our current experimental the training data is randomly shuffled before every epoch of RMSProp based training. Dropout regularization \cite{bhattacharya2008handwritten} is used (only in the FC layers) to reduce the possibility of overfitting of the network during training The parameters of the learning algorithm used to train the proposed architecture is presented in \ref{tab:Train_Parameters}.

\begin{equation} \label{Eq:CEL}
H(y,\hat{y}) = -\sum_{i=1}^{k}[y_{i}\ln{\hat{y_{i}}}]
\end{equation}
where, $y$ is actual class, $\hat{y}$ is predicted class, $k$ is total number of classes.
\newline
\newline

During training the individual are decoded for loading the kernels to every layer of each column. In our three-column based architecture as there are three layers at every column, the first three elements (kernel values) are loaded to three layers of first column respectively. Similarly, next three elements are loaded to the layers of second column and last three kernel values are loaded to the layers of the third column.

After training the best fit individual is selected from the population and update the kernel sizes of every layers of all the columns accordingly. The PyTorch code is implemented in the following link: \url{https://github.com/DeepQn/GA-Based-Kernel-Size}.

\section{Experiments} \label{sec:EX}
As mentioned earlier, a genetic algorithm based approach is proposed for recognizing handwritten character and digit images of multiple Bangla handwritten scripts. A Python based library, PyTorch is used to implement, train and test the proposed multi-column CNN based architecture. MATLAB is used to perform the basic image processing operations. All of the experiments are performed using systems with Intel dual core i$5$ processors, $8$ GB RAM and a NVIDIA GeForce $1050$ Ti graphics card with $4$ GB internal memory.

\subsection{Datasets used in our experiment}
The proposed architecture is tested on three publicly available multiple datasets of Bangla scripts. The name, type and volume (number of training samples and testing samples) of the datasets used in our current experimental setup are given in \ref{tab:Dataset}. These intricate handwritten datasets are significantly different from each other, thus while considering these multiple datasets constitute an ideal test set for our proposed architecture. More information about the datasets is given in the last column of \ref{tab:Dataset}.

\subsection{Pre-processing of datasets}
A few numbers of pre-processing steps are used to process the images of the datasets used. Every image on isolated handwritten characters or digits is binarized and centre cropped by the tightest bounding box \cite{sarkhel2017multi} and finally resized to 32 x 32 pixels. Median and Gaussian filters are used to remove noises from the images.

After the pre-processing step, the training dataset is randomly divided into training set and validation set in such a way that the size of the validation set matches with the size of the test set. Now, the architecture is trained on the training set and saved against the best recognition accuracy achieved on validation set. After the network is trained, it is evaluated on the test set.

\begin{small}
\begin{table*}
\caption{A comparative analysis of the proposed methodology with some of the popular contemporary works.} \label{tab:Result_Contemp}
\vspace{0.3cm}
\centering
\resizebox{\linewidth}{!}{%
\begin{tabular}{c c c}
\hline
Dataset type & Work reference & Recognition accuracy (\%) \\
\hline
ISI digits & Sharif et al. \unskip~\cite{sharif2016hybrid} & 99.02 \% \\
                & Wen et al. \unskip~\cite{wen2012classifier} & 96.91 \% \\
                & Das et al. \unskip~\cite{das2012genetic} & 97.70 \% \\
                & Akhnad et al. \unskip~\cite{rahman2015bangla} & 97.93 \% \\
                & CNNAP \unskip~\cite{akhand2016convolutional} & 98.98 \% \\
                & The Present Work & \textbf{99.12} \% \\

\hline
Bangla basic characters & Roy et al. \unskip~\cite{roy2012region} & 86.40 \% \\
                        & Das et al. \unskip~\cite{das2010handwritten} & 80.50 \% \\
                        & Basu et al. \unskip~\cite{basu2009hierarchical} & 80.58 \% \\
                        & Sarkhel et al. \unskip~\cite{sarkhel2015enhanced} & 86.53 \% \\
                        & Bhattacharya et al. \unskip~\cite{bhattacharya2006recognition} & 92.15 \% \\
                        & Lecun et al. \unskip~\cite{le1994word} & 92.88 \% \\
                        & The Present Work & \textbf{97.10} \% \\
\hline
Bangla compound characters & Das et al. \unskip~\cite{das2010handwritten} & 75.05 \% \\
                           & Das et al. \unskip~\cite{das2015handwritten} & 87.50 \% \\
                           & Sarkhel et al. \unskip~\cite{sarkhel2015enhanced} & 78.38 \% \\
                           & Sarkhel et al. \unskip~\cite{sarkhel2016multi} & 86.64 \% \\
                           & Lecun et al. \unskip~\cite{le1994word} & 86.85 \% \\
                           & Roy et al . \unskip~\cite{roy2017handwritten} & 90.33 \% \\
                           & The Present Work & \textbf{94.77} \% \\
\hline
\end{tabular}%
}
\end{table*}
\end{small}

\subsection{Experimental results}
As mentioned before, the proposed technique is tested on three Bangla handwritten datasets. Among these three, one is Bangla digits dataset, one is Bangla basic characters dataset and the last one is Bangla compound characters dataset. The best recognition accuracy achieved on these three different datasets using our proposed methodology is presented in \ref{tab:Results}. For comparison, the recognition accuracies on these datasets using our proposed architecture on fixed scale i.e. in every layer using kernel size of only $3 \times 3$ are presented in \ref{tab:Results3}, kernel size of only $5 \times 5$ are presented in \ref{tab:Results5} and kernel size of only $7 \times 7$ are presented in \ref{tab:Results7}. From the experimental results we found that the network using kernel size $5 \times 5$ gives better performance than the network using kernel size $3 \times 3$ or $7 \times 7$. In case of the fixed kernel size the performance first increases from kernel size $3 \times 3$ to kernel size $5 \times 5$ and then again reduces for the kernel size $7 \times 7$. In case of multi-scaling, the network gives a better performance than all of these three fixed-scaled kernels. However, all the combinations which result in multi-scaling may not give better recognition accuracy than a fixed-scale kernel. This optimal combination of multi-scaling can be found using a genetic algorithm based approach which is the primary concern of this work.

To prove the efficiency of our proposed work, in \ref{tab:Result_Contemp} we have presented the recognition performance of some of the contemporary works on the datasets used in our current experimental setup. The best recognition accuracy achieved by a system is made boldface in \ref{tab:Result_Contemp}.

\section{Conclusion} \label{sec:CO}
In our present work a methodology has been proposed to reduce the effort of finding the optimal combination of kernel sizes for the layers of a neural network based architecture. The genetic algorithm based technique selects the optimal kernel combination from an initial population of different kernel sizes after iterating over multiple generations. The researchers in the field of computer vision and other domains also can utilize this neural architecture search strategy to initialize optimal combination of kernel sizes including other hyper-parameters of their neural network based architectures before final training. This will results better performance along with less human effort. This genetic algorithm based neural architecture search methodology opens a new area of research towards pattern recognition including other domains.

\section*{Acknowledgments}
The authors are thankful to the Center for Microprocessor Application for Training Education and Research (CMATER) and Project on Storage Retrieval and Understanding of Video for Multi- media (SRUVM) of Computer Science and Engineering Department, Jadavpur University, for providing infrastructure facilities during progress of the work. The current work, reported here, has been partially funded by University with Potential for Excellence (UPE), Phase-II, UGC, Government of India.

%
%
%
%
%

\bibliographystyle{splncs04}

\bibliography{References}
\end{document}